\documentclass{article}

\usepackage{arxiv}

\usepackage[utf8]{inputenc} 
\usepackage[T1]{fontenc}    
\usepackage{hyperref}       
\usepackage{url}            
\usepackage{booktabs}       
\usepackage{amsfonts}       
\usepackage{nicefrac}       
\usepackage{microtype}      
\usepackage{lipsum}
\usepackage{graphicx}
\usepackage{multirow}
\usepackage{color}
\usepackage{soul}
\usepackage[square,numbers,sectionbib]{natbib}
\usepackage{cleveref}
\graphicspath{ {./images/} }

\usepackage{soul}
\usepackage{makecell}
\usepackage{siunitx}
\usepackage{fontawesome}
\usepackage{xcolor} 

\title{Extracting and filtering paraphrases by bridging natural language inference and paraphrasing}

\author{
  Matej Klemen \\
   University of Ljubljana \\
   Faculty of Computer and Information Science \\
   Večna pot 113, 1000 Ljubljana, Slovenia \\
  \texttt{matej.klemen@fri.uni-lj.si} \\
   \And
 Marko Robnik-Šikonja \\
   University of Ljubljana \\
   Faculty of Computer and Information Science \\
   Večna pot 113, 1000 Ljubljana, Slovenia \\
  \texttt{marko.robnik@fri.uni-lj.si} \\
}

\begin{document}
\date{}
\maketitle
\begin{abstract}
Paraphrasing is a useful natural language processing task that can contribute to more diverse generated or translated texts. Natural language inference (NLI) and paraphrasing share some similarities and can benefit from a joint approach. We propose a novel methodology for the extraction of paraphrasing datasets from NLI datasets and cleaning existing paraphrasing datasets. Our approach is based on bidirectional entailment; namely, if two sentences can be mutually entailed, they are paraphrases. We evaluate our approach using several large pretrained transformer language models in the monolingual and cross-lingual setting. The results show high quality of extracted paraphrasing datasets and surprisingly high noise levels in two existing paraphrasing datasets.
\end{abstract}


\section{Introduction}
Natural language inference (NLI) and text paraphrasing are two useful and frequently applied tasks in natural language processing.
In NLI, the task is to determine if a target sequence (hypothesis) entails the source sequence (premise), contradicts it, or is neutral with respect to it.
A paraphrase is a restatement of the meaning of a text or passage using other words.
Both tasks are commonly used, either on their own or to support other tasks. 
For example, NLI can be used for fact verification, as demonstrated in the Fact Extraction and Verification 2018 Shared Task \cite{thorne-etal-2018-fever}, while paraphrasing can be used to rephrase news headlines~\cite{paraphrasing-headlines} or improve text expression in grammar tools. 

Despite mainly being researched separately, NLI and paraphrasing are similar and can benefit from a joint approach. As we show, the performance of paraphrasing can be improved by using additional resources available for NLI. For example, paraphrase generation methods can generate more diverse paraphrases by being trained using paraphrases extracted from existing NLI datasets. 
In the situation where little or no resources are available to train a paraphrase identification model, a pretrained NLI model could be used as an alternative.

Although large English datasets exist for both NLI and paraphrasing, paraphrasing datasets are often automatically derived from datasets containing similar texts but not necessarily paraphrases. For example, COCO captions dataset \cite{chen2015microsoft}) contains multiple captions of the same image, and Quora question pairs are composed of duplicate questions. We aim to extend the pool of available paraphrasing resources using a framework whose primary use is the extraction of paraphrases from existing NLI datasets.

In the proposed framework, we treat paraphrases as sequence pairs where the hypothesis entails the premise and the premise entails the hypothesis, i.e. there is a bidirectional entailment. This connection has previously been mentioned \cite{paraphrasing-nli-survey-2010}, but its potential for paraphrase extraction and filtering was not fully explored.
We study the following two scenarios: 
\begin{enumerate}
    \item \textbf{Creation of a paraphrase dataset from an existing NLI dataset.} We demonstrate this scenario in a monolingual and a cross-lingual setting, using Stanford NLI (SNLI) \cite{bowman-etal-2015-snli}, Multi-genre NLI (MNLI) \cite{williams-etal-2018-mnli} and Cross-lingual NLI (XNLI) \cite{conneau-etal-2018-xnli} datasets to produce new paraphrasing datasets. In the cross-lingual setting, we compare extracted paraphrases with the so-called ``translate-train'' and ``translate-test'' methods, which transform the cross-lingual into a multilingual problem by automatically translating the training or the test set.
   
    \item \textbf{Cleaning of an existing paraphrasing dataset.} By a slight modification, we reuse the framework to filter false positive instances, i.e. instances which are labeled as paraphrases but are not actual paraphrases. The filtering uses the validation with the  bidirectional entailment relation. We demonstrate this scenario on two English paraphrase identification datasets: Quora Question Pairs (QQP) and Microsoft Research Paraphrase Corpus (MRPC) \cite{dolan-brockett-2005-mrpc}. 
\end{enumerate}

In our approach, we compare several NLI classifiers with varying performance. 
To show the usability for different use-cases that might have different expectations concerning the correctness of paraphrases, we present the results at multiple certainty levels and show that precision and recall can be used to select the right setting. For example, when building a model for rephrasing news, a more  ``conservative'' paraphrase dataset might be preferable in order to avoid a model that would make up facts.
To facilitate further research, we release the source code of our framework \footnote{\url{https://github.com/matejklemen/paraphrase-nli}}.

The paper is split into further four sections. In \Cref{sec:related-work}, we present a related work on paraphrasing and NLI. In \Cref{sec:methodology},
we present our methodology for creation and cleaning of new paraphrasing datasets. In \Cref{sec:evaluation}, we evaluate the newly proposed methods. Finally, in \Cref{sec:conclusions}, we summarize the work and give directions for further work.

\section{Related work}
\label{sec:related-work}

We first overview the related work that connects NLI with other tasks, followed by existing work on the construction of paraphrasing datasets.

\subsection{Connections between NLI and other tasks}

Since the early days of research on textual entailment, various authors have used it to improve other tasks' performance. For example, \citet{lloret-etal-2008-entail-summary} use the textual entailment recognition to improve a summarization system. In order to generate preliminary summaries, they iterate through sentences and only keep those that did not entail any other previously seen sentences.
To improve an information retrieval system, \mbox{\citet{dekang-pantel-dirt}} extract textual entailment rules from text by using similarities in dependency paths.
\citet{harabagiu-hickl-2006-methods} integrate textual entailment into the open-domain question answering task, observing improved performance when using it to filter answer candidates or as a ranking mechanism for passages.
Similarly, \mbox{\citet{chen-etal-2021-nli-models}} use textual entailment to verify that the answer to the question entails a given context.
The entailment was also used to improve the evaluation procedure of other tasks. \citet{pado-etal-2009-robust} propose a metric for machine translation evaluation that incorporates bidirectional textual entailment as a way to measure meaning equivalence between a hypothesis and a reference summary. In our work, we use the bidirectional textual entailment for three purposes. First, to create new paraphrasing datasets.  Second, we show the feasibility of the idea in the cross-lingual setting, and third, we show that the idea can be reused for filtering existing paraphrasing datasets. 

With the introduction of large annotated datasets for many tasks, authors started to use them in transfer learning. For example, \citet{white-etal-2017-inference} convert semantic proto-role labeling, paraphrasing and anaphora resolution datasets into a binary textual entailment dataset to explore the extent to which an NLI model can capture specific aspects of semantics. Two further examples of such datasets (QNLI and WNLI) are included in the GLUE benchmark (General Language Understanding Evaluation) \cite{wang-etal-2018-glue}. QNLI is derived from the Stanford question answering dataset \cite{rajpurkar-etal-2016-squad} that contains question-context pairs, where one of the sentences in the context contains the answer to the question. To produce QNLI, each pair is converted into multiple question-sentence pairs labeled as entailment if that sentence contains the answer to the question and non-entailment otherwise.
WNLI is derived from the Winograd schema challenge dataset \cite{levesque-etal-2012-wsc} that contains sentences involving an ambiguous pronoun and a list of possible referents of that pronoun. To produce WNLI, the pronoun is replaced by each possible referent and labeled as entailment if that referent refers to the pronoun.

\subsection{Existing paraphrasing datasets}
Existing paraphrasing datasets have been created (or repurposed) using a variety of data sources including image captions - COCO \cite{chen2015microsoft} and Multi30k \cite{elliott-etal-2016-multi30k}), duplicate questions - QQP and PAWS\footnote{PAWS also contains instances from Wikipedia as an additional source.} \cite{zhang-etal-2019-paws}), news data - MRPC \cite{dolan-brockett-2005-mrpc}), tweets - Twitter News URL corpus \cite{lan-etal-2017-continuously} and Twitter Paraphrase corpus \cite{xu-etal-2014-extracting}), subtitles - Opusparcus \cite{creutz-2018-open}), and more general bilingual text - PPDB \cite{ganitkevitch-etal-2013-ppdb}).
Although most resources are available for the English language, some datasets also exist for other languages. 
Opusparcus uses multilingual subtitles to produce paraphrases in six different languages. PPDB was created for English and Spanish, and PAWS-X \cite{yang-etal-2019-paws} contains instances from PAWS translated into six different languages.
The Multi30k dataset provides a translation of the image captions in Flickr30k \cite{young-etal-2014-image} to German, to which they add new (independent) German captions.

With our work, we expand the pool of paraphrasing resources by using a different, previously unused source -- NLI data. 
By learning the relations in NLI datasets and asserting the validity of bidirectional entailment for paraphrasing, we ensure the equivalence of meanings. In some of the existing paraphrasing datasets, the meaning equivalence can be potentially problematic, e.g., in those that use image captions or duplicate questions. However, a slight modification of our framework enables us to detect some of the inconsistencies in existing paraphrasing datasets.

Our approach can be used in any language with an existing NLI dataset and NLI model of sufficient quality. We verify this with cross-lingual experiments where we extract paraphrases for a few less-resourced languages using a cross-lingual transfer from the English NLI model.

\section{Transfer between NLI and paraphrasing datasets}
\label{sec:methodology}
In this section, we describe the use of NLI for extracting paraphrases and paraphrase filtering. In \Cref{sec:paraphrase-extraction}, we first describe the methodology for extraction of paraphrases from an existing NLI dataset, while in \Cref{sec:paraphrase-filtering}, we describe how an existing paraphrasing dataset can be cleaned.
In \Cref{sec:implementation-details}, we discuss the training of NLI classifiers which are the essential components of our methodology.

\begin{figure}[tbh]
    \centering
    \includegraphics[width=\columnwidth]{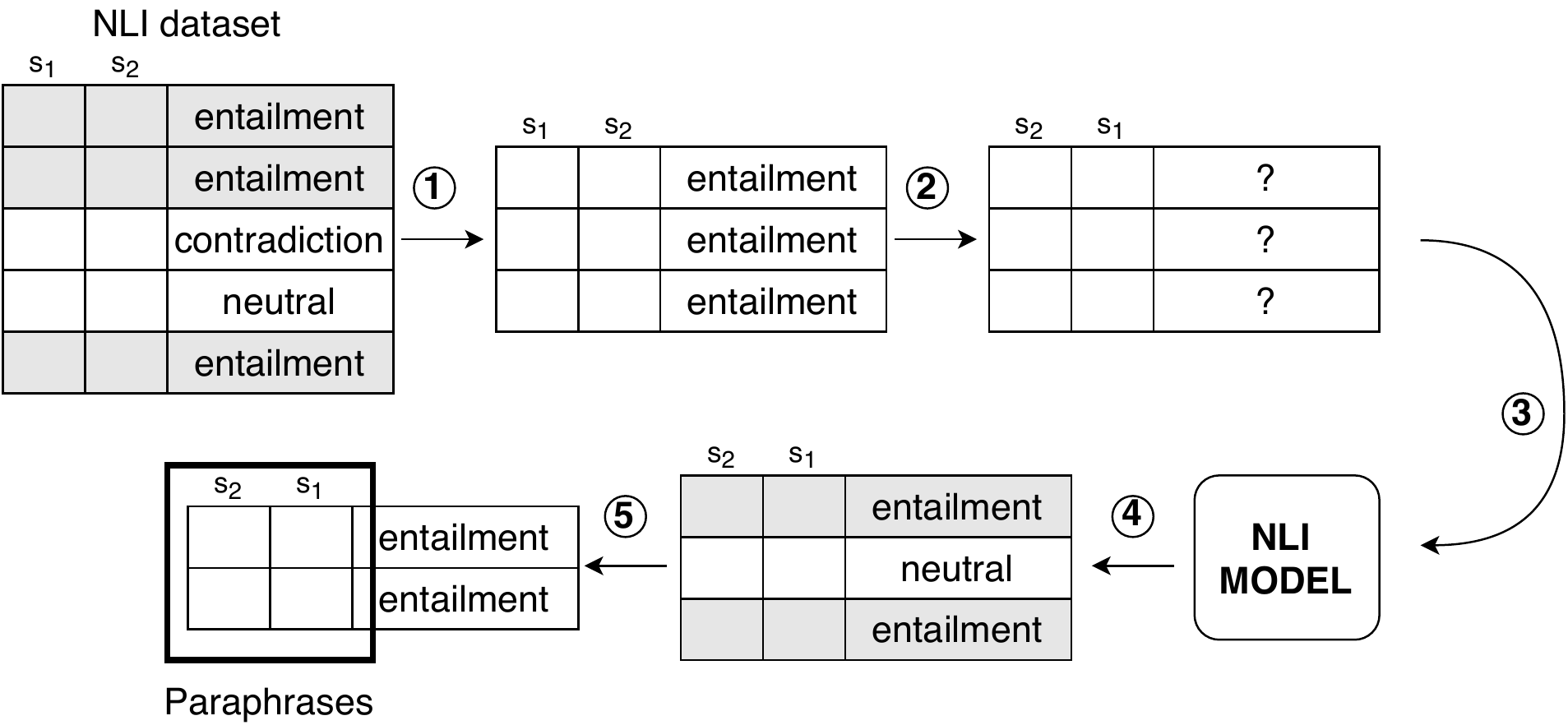}
    \caption{A schematic representation of our method for paraphrase extraction from an NLI dataset. In Step 1, we extract the sequence pairs labeled as entailment from the NLI dataset. In Step 2, we reverse the order of these sequence pairs. The labels for the reversed instances are unknown, so we use an NLI model to predict them (Steps 3 and 4). Once we obtain the new labels, we select the sequence pairs that are marked as entailment (Step 5). These make up the new paraphrasing dataset.}
    \label{fig:paraphrase-extraction}
\end{figure}

\subsection{NLI for paraphrase extraction}
\label{sec:paraphrase-extraction}
Figure~\ref{fig:paraphrase-extraction} shows five steps of the proposed methodology for extraction of paraphrases from an existing NLI dataset. The main idea of the methodology is that two sentences forming a paraphrase have to mutually entail each other in the logical sense.
The input to the methodology is a high quality (e.g., manually labeled) NLI dataset, where instances are pairs of sentences $(s_1, s_2)$, i.e. (premise, hypothesis). In NLI datasets, there are typically three types of relations between the sentences $s_1$ and $s_2$, namely entailment, contradiction, or neutral.

In Step 1, we filter the input NLI dataset, keeping only the instances labeled as the entailment. 
In Step 2, we reverse the order of these sequences, placing each hypothesis in place of the premise and the premise in place of the hypothesis. We obtain a new NLI dataset with unknown labels, so in Step 3, we use an existing NLI prediction model to predict them. This can be any NLI model; for example, we can obtain it by training a machine learning classifier on the original NLI dataset or some other NLI dataset.

The result (Step 4) are pairs of sentences, labeled as entailment, contradiction or neutral by the trained NLI model.
We filter the results (Step 5), keeping only the instances predicted to be entailment. 
For these instances, the bidirectional textual entailment relation holds and, therefore, they constitute paraphrases. 
To form the final paraphrasing dataset, we remove the labels, keeping only sentence pairs $(s_2, s_1)$.

The instances which are not extracted are not paraphrases and could be used as negative instances in a paraphrase detection or generating dataset. As our experiments show, the number of non-paraphrases is much larger than the number of paraphrases. They range from trivial to difficult-to-detect non-paraphrases. In order to find difficult instances, Step 5 can be modified only to keep the instances predicted to be neutral. In our experiments, we only deal with the extraction of paraphrases and leave the detailed exploration of this scenario for further work.

\subsection{NLI for paraphrase filtering}
\label{sec:paraphrase-filtering}

The above method for paraphrase extraction can be reused for other tasks. We describe how it can be adapted to filter an existing paraphrase identification dataset, where examples are labeled as paraphrases or non-paraphrases. Such datasets may contain a certain amount of noisy annotations. For instance, in the QQP dataset, noisy instances are present because a question and its duplicate question (a paraphrase) are usually, but not always, identical concepts. The authors motivate duplicate question detection as finding questions with the same intent.
For example, ``How do I learn SQL?'' and ``Which is the best book for SQL?'' would indicate the exact intent of obtaining more knowledge about SQL, but the meaning cannot be considered identical.

To use the framework for paraphrase filtering, we make the following modifications.
The input to our framework is now a paraphrase identification dataset, with the labels paraphrase and non-paraphrase.
In Step 1, we filter the dataset by selecting instances labeled as a paraphrase (instead of entailment).
We repeat the process of reversing the order of paraphrases (Step 2), now predicting if the reversed sequence pair is a paraphrase or non-paraphrase using an existing paraphrase identification model (Step 3 and 4).
In Step 5, we select the instances predicted to be non-paraphrases. These instances are paraphrases in one direction but not in the other. Since a paraphrase is supposed to be symmetrical, the output represents falsely labeled paraphrases removed in the final cleaned dataset.

\subsection{Implementation details}
\label{sec:implementation-details}
Sections \ref{sec:paraphrase-extraction} and \ref{sec:paraphrase-filtering} presents the description of the proposed method.  Here, we provide implementation details. For all NLI models, we use  the default sequence pair set up \cite{devlin-etal-2019-bert}. 

In our fine-tuning procedure, we consider four types of models: BERT, multilingual BERT \cite{devlin-etal-2019-bert}, RoBERTa (base and large variant) \cite{liu-2019-roberta}, and XLM-RoBERTa (XLM-R, base and large variant) \cite{conneau-etal-2020-unsupervised}.
We fine-tune them on the datasets described in \Cref{sec:datasets}, 
considering multilingual BERT and XLM-R in cross-lingual experiments and BERT and RoBERTa in monolingual experiments. In Section \ref{sec:evaluation}, we report results for only one of the two model types, selected based on the validation set $F_1$ score.
We use the AdamW optimizer \cite{loshchilov-2019-adamw} with the learning rate \num{2e-5}. In one instance, where the model did not converge (large XLM-R on XNLI), we decreased the learning rate to \num{5e-6}.
We try two settings of the maximum sequence length for each model: $95$th and $99$th percentile of sequence lengths in a training set.
To prevent overfitting, we use early stopping based on the validation set $F_1$ score using a tolerance of five rounds.
We use the model implementations from the transformers package \cite{wolf-etal-2020-transformers} and pretrained checkpoints from the HuggingFace Model Hub\footnote{\url{https://huggingface.co/models}}. 

\section{Evaluation}
\label{sec:evaluation}
The proposed methodology can create paraphrasing datasets from NLI datasets or improve the quality of existing paraphrasing datasets.  We first describe the NLI and paraphrasing datasets used as inputs to our methodology (\Cref{sec:datasets}), and the metrics we use to measure the performance of the models (\Cref{sec:metrics}). In Section \ref{sec:extraction-results} we evaluate the creation of paraphrasing datasets, and in Section \ref{sec:filtering-results}, we analyse the cleaning of existing paraphrasing datasets.

\subsection{Input NLI and paraphrasing datasets}
\label{sec:datasets}
In our experiments, we use three NLI datasets (SNLI, MNLI, and XNLI) and two paraphrase identification datasets (QQP and MRPC).
All instances in NLI datasets are annotated using a three-class annotation scheme where each sequence pair is labeled as entailment, neutral, or contradiction. The paraphrase identification datasets are annotated using two labels: paraphrase or non-paraphrase.
The NLI datasets are balanced in the distributions of the three labels. In contrast, the paraphrase identification datasets are unbalanced, with MRPC containing approximately two-thirds of paraphrases and QQP approximately two-thirds of non-paraphrases. Unless described otherwise, we use the training:development:testing set splits determined by dataset authors.

\textbf{SNLI} \cite{bowman-etal-2015-snli} is a NLI dataset containing \num{569033} instances. The premises are image captions taken from the Flickr30k corpus \cite{young-etal-2014-image}. The hypotheses were produced by human contributors asked to write one caption that is definitely true (entailment), one that might be true (neutral) and one that is definitely not true (contradiction) given the premise. The same meaning of labels is also used in MNLI and XNLI.

\textbf{MNLI} \cite{williams-etal-2018-mnli} is an NLI dataset containing \num{412349} instances that include written and spoken English across ten genres, as opposed to the single genre present in SNLI. The premises are obtained from ten sources of freely available text while obtaining hypotheses is the same as for SNLI. We use a slightly different dataset split from the one proposed by the authors.
The validation and test sets in the original version are split into two parts, one containing genres present in the training set and the other containing genres not present in the training set. Models are usually evaluated separately on the two parts, but we combine the two parts in our experiments.
The test set labels are private, so we cannot use them. Instead, we split the combined validation set randomly into two equal parts and use one half as the validation and one half as the test set.

\textbf{XNLI} \cite{conneau-etal-2018-xnli} is a NLI dataset containing \num{112500} instances divided evenly across $15$ languages. In contrast to the previous two NLI datasets, XNLI only contains the validation and test set, while the training set is reused from MNLI.
The sequence pairs were obtained using the same procedure as in MNLI and human translated into $15$ languages. The authors also provide a machine translation of the training set into $15$ languages and a machine translation of the validation and test sets into English, which we use to train monolingual models showing upper bounds of performance.

\textbf{QQP}\footnote{\url{https://quoradata.quora.com/First-Quora-Dataset-Release-Question-Pairs}} is a paraphrase identification dataset containing \num{404279} instances. It contains potentially duplicate question pairs posted on the Quora question-and-answer platform. The labels indicate whether a pair of questions could be considered a duplicate or not. 
We use the dataset version provided in the GLUE benchmark \cite{wang-etal-2018-glue}. As the test set labels are private,  we use a randomly selected 50\% of the validation set as the test set instead.

\textbf{MRPC} \cite{dolan-brockett-2005-mrpc} is a paraphrase identification dataset containing \num{5801} instances. Its construction consisted of an initial candidate selection, noisy paraphrase detection using a machine learning model with handcrafted features, and human annotation of a subset of detected paraphrases. As there is no pre-determined validation set, we constructed it from \num{1000} instances, randomly sampled from the training set.

\subsection{Evaluation metrics}
\label{sec:metrics}

To assess the properties of our method, we use several metrics on the trained NLI models and also perform a human evaluation of a sample of the instances.

For the NLI models, we measure the (binary) precision (P) and recall (R) for the label entailment.
These provide us with an estimate of how many and how correct paraphrases NLI models are able to extract: high precision indicates many correct paraphrases, and high recall indicates good coverage of paraphrases existing in the NLI dataset.
P and R are defined with Equation \ref{eqn:recall}, where $tp$ is the number of actual entailment instances that are correctly predicted by the model, $fp$ the number of actual non-entailment instances that are incorrectly predicted by the model, and $fn$ the number of actual entailment instances that are incorrectly predicted by the model.
\noindent\begin{minipage}[b]{.5\linewidth}
$$ 
\label{eqn:precision}
P = \frac{tp}{tp + fp} \nonumber
$$ 
\end{minipage}%
\begin{minipage}[b]{.5\linewidth}
 \begin{equation}
 \label{eqn:recall}
R = \frac{tp}{tp + fn}
\end{equation}
\end{minipage}
\vspace{1mm}

Using P and R, we rely on the NLI model performance as a reasonable proxy for paraphrase extraction.
To validate these numbers, we perform a small scale manual validation. We randomly sample $100$ extracted paraphrases for each setting and check its precision (to compute recall, we would have to check the entire NLI dataset). On XNLI, which consists of 15 languages, we manually validated German and French results as experts in other languages were not accessible to us. In manual validation, we mark as paraphrases only those sequence pairs, for which both sequences contain exactly the same meaning.

For the evaluation of paraphrase filtering, we use the same metrics, but we take non-paraphrase as the positive label.

\subsection{Paraphrase extraction results}
\label{sec:extraction-results}
Using the methodology described in \Cref{sec:paraphrase-extraction}, we processed the three NLI datasets (SNLI, MNLI, and XNLI) described in \Cref{sec:datasets}. We first present the monolingual paraphrase extraction, followed by the cross-lingual experiments.

\subsubsection{Monolingual paraphrase extraction}
Table \ref{tab:results-snli-mnli} presents the number of extracted paraphrases, precision and recall of the used NLI model, and a manual estimate of precision in two monolingual datasets, SNLI and MNLI.  We show results for three decision boundaries: when the entailment class is the most likely of the three (argmax) and two stricter probability thresholds for the entailment class ($0.75$ and $0.90$).

\begin{table*}[htb]
\caption{The number of extracted paraphrases (\#), precision (P), recall (R) and manually estimated precision (\faHandPaperO P) in SNLI and MNLI using three different decision thresholds for the entailment label. Below each dataset name we show the number of entailment instances in the dataset.} 
\centering
\resizebox{0.75 \linewidth}{!}{
\begin{tabular}{cl@{\hskip 30pt}cccc@{\hskip 30pt}cccc}
\toprule
dataset & decision & \multicolumn{4}{c}{roberta-base} & \multicolumn{4}c{roberta-large} \\
(size) & threshold & \# & P & R & \faHandPaperO P & \# & P & R & \faHandPaperO P \\
\midrule
\multirow{3}{*}{\makecell{SNLI \\ ($183416$)}}
& argmax    & $17793$ & $0.902$ & $0.901$   &  $0.670$ & $16300$   & $0.921$ & $0.906$ & $0.790$ \\
& T=$0.75$  & $12067$ &  $0.948$ & $0.825$ & $0.790$ & $10339$ & $0.961$ & $0.817$ & $0.900$ \\
& T=$0.90$  & $8857$ &  $0.972$ & $0.730$ & $0.850$ & $4539$ & $0.985$ & $0.547$ & $0.920$ \\
\midrule
\multirow{3}{*}{\makecell{MNLI \\ ($130899$)}}
& argmax  & $40948$ &  $0.888$ & $0.865$ & $0.820$ & $43992$ & $0.909$ & $0.886$ & $0.860$ \\
& T=$0.75$  & $30718$ &  $0.945$ & $0.734$ & $0.930$ & $35789$ & $0.956$ & $0.799$ & $0.910$ \\
& T=$0.90$  & $21905$ &  $0.975$ & $0.576$ & $0.950$ & $27119$ & $0.976$ & $0.661$ & $0.950$ \\

\bottomrule
\end{tabular}
}
\label{tab:results-snli-mnli}
\end{table*}

We notice that the proposed approach extracts a significant number of paraphrases in all tested modes. 
Below we address several aspects of the obtained paraphrases.

First, although SNLI contains a larger candidate pool of potential paraphrases than MNLI, there is a larger proportion of paraphrases present in MNLI. A likely reason for this is heuristics employed by the MNLI annotators to generate hypotheses quickly; some have previously been outlined for SNLI and MNLI by \citet{gururangan-etal-2018-annotation}.
MNLI contains multiple genres, for some of which the quickest way to create a hypothesis given a premise is to rephrase the premise.
This explanation is strengthened by the genre distribution of extracted paraphrases, with the most common genres being telephone and fiction.
Premises from these two genres are often parts of dialogues. It is difficult for annotators to quickly find a true hypothesis without it being a rephrased version of the premise. 

Second, using a larger and more precise roberta-large model does not necessarily imply a higher amount of extracted paraphrases. For example, in MNLI, the large model extracts more paraphrases than the base model, but in SNLI, this is not the case.
When using the two stricter decision thresholds, we can see the reason: the larger model obtains higher precision, but lower recall, which roughly corresponds to the decrease in the number of extracted paraphrases. On the other hand, the explanation for the argmax decision boundary is less clear.
We checked the rate and type of errors the two models make and found that the base model more frequently incorrectly classifies entailment as contradiction than the large model ($80$ versus $38$ errors). This could be an indication that the model has learned an erroneous pattern which leads to the model extracting more false positives. The manual validation confirmed that the base model indeed extracts less correct paraphrases.

Unsurprisingly, with more strict decision thresholds, we extract fewer paraphrases. Different thresholds represent trade-offs between the precision and recall of the paraphrase extraction process. Higher thresholds lead to the extraction of paraphrases with higher confidence. Thus, using higher thresholds, we decrease the likelihood of extracting false paraphrases and increase the chance that we miss some of the actual paraphrases. 
The results of the manual validation agree with this, although the improvement in precision becomes smaller as we use stricter decision thresholds.
More certain paraphrases are typically more conservative. Our additional analysis showed that the sequences in more certain pairs are more similar in length (number of tokens) and have a smaller normalized edit distance \cite{levenshtein1966bcc} than the paraphrases extracted with the argmax decision boundary. 
Stricter decision boundaries might be preferred when we want to minimize the false positives in paraphrases that could guide the downstream paraphrase generation model to make up new information.  

\subsubsection{Cross-lingual paraphrase extraction}
Table \ref{tab:results-xnli} shows the number of extracted paraphrases in the cross-lingual setting using the XNLI datasets in $15$ languages. We train and validate a single model on the English data and use this model on the validation and test sets in all $15$ languages. We report only the results obtained with the argmax decision boundary, i.e. we extract a paraphrase if entailment is the most probable class. The other two decision thresholds extract lower numbers of paraphrases in similar ratios as shown in Table \ref{tab:results-snli-mnli}. 

\begin{table*}[htb]
\caption{The number of extracted paraphrases (\#), precision (P) and recall (R) for datasets in 15 different languages present in XNLI for the argmax decision threshold. Each dataset contained \num{2500} entailment instances. We mark machine translation (MT) settings where a language-specific model could not be found or set up with ``/''. 
For German and French, we  provide a human estimate of the precision (\faHandPaperO P).}
\centering
\resizebox{0.75 \linewidth}{!}{
\begin{tabular}{
l@{\hskip 15pt}
c@{\hskip 5pt}c@{\hskip 5pt}c@{\hskip 25pt}
c@{\hskip 5pt}c@{\hskip 5pt}c@{\hskip 5pt}c@{\hskip 25pt} 
c@{\hskip 5pt}c@{\hskip 5pt}c@{\hskip 25pt}
c@{\hskip 5pt}c@{\hskip 5pt}c}
\toprule
lng & \multicolumn{3}{c}{XLM-R base} & \multicolumn{4}{c}{XLM-R large} & \multicolumn{3}{c}{MT (train)} & \multicolumn{3}{c}{MT (test)} \\
 & \# & P & R & \# & P & R & \faHandPaperO P & \# & P & R & \# & P & R \\
\midrule
ar  & $405$ & $0.768$ & $0.587$ & $394$ & $0.866$ & $0.629$ & / & $624$ & $0.761$ & $0.710$ & $331$ & $0.900$ & $0.562$ \\
bg  & $449$ & $0.808$ & $0.686$ & $439$ & $0.882$ & $0.735$ & / & $452$ & $0.799$ & $0.722$ & $388$ & $0.889$ & $0.670$ \\
de  & $494$ & $0.783$ & $0.686$ & $465$ & $0.872$ & $0.740$ & $0.780$ & $560$ & $0.830$ & $0.792$ & $399$ & $0.880$ & $0.660$ \\
el  & $488$ & $0.777$ & $0.667$ & $446$ & $0.860$ & $0.726$ & / & $431$ & $0.813$ & $0.670$ & $382$ & $0.895$ & $0.655$ \\
en  & $488$ & $0.831$ & $0.795$ & $509$ & $0.890$ & $0.843$ & / & $521$ & $0.907$ & $0.874$ & $521$ & $0.907$ & $0.874$ \\
es  & $462$ & $0.804$ & $0.723$ & $462$ & $0.893$ & $0.740$ & / & $505$ & $0.811$ & $0.774$ & $393$ & $0.896$ & $0.705$ \\
fr  & $512$ & $0.780$ & $0.737$ & $460$ & $0.881$ & $0.741$ & $0.810$ & $480$ & $0.875$ & $0.791$ & $386$ & $0.884$ & $0.692$ \\
hi  & $383$ & $0.746$ & $0.532$ & $383$ & $0.841$ & $0.607$ & / & $620$ & $0.575$ & $0.644$ & $268$ & $0.847$ & $0.446$ \\
ru  & $438$ & $0.781$ & $0.665$ & $411$ & $0.882$ & $0.663$ & / & $465$ & $0.784$ & $0.703$ & $356$ & $0.884$ & $0.611$  \\
sw  & $407$ & $0.657$ & $0.580$ & $371$ & $0.812$ & $0.560$ & / & / & / & / & $242$ & $0.841$ & $0.380$  \\
th  & $381$ & $0.815$ & $0.544$ & $387$ & $0.851$ & $0.644$ & / & / & / & / & $302$ & $0.867$ & $0.499$ \\
tr  & $405$ & $0.764$ & $0.619$ & $367$ & $0.865$ & $0.662$ & / & $466$ & $0.793$ & $0.724$ & $331$ & $0.878$ & $0.569$ \\
ur  & $307$ & $0.760$ & $0.449$ & $358$ & $0.823$ & $0.531$ & / & / & / & / & $244$ & $0.837$ & $0.384$ \\
vi  & $374$ & $0.785$ & $0.649$ & $380$ & $0.863$ & $0.650$ & / & $563$ & $0.762$ & $0.645$ & $278$ & $0.866$ & $0.531$ \\
zh  & $419$ & $0.784$ & $0.610$ & $407$  & $0.854$ & $0.640$ & / & $597$ & $0.778$ & $0.736$ & $349$ & $0.893$ & $0.567$ \\ 
\bottomrule
\end{tabular}
}
\label{tab:results-xnli}
\end{table*}

The results indicate that the proposed approach is effective even in the cross-lingual setting and produces valuable paraphrasing datasets. Below we analyze their properties.

While the number of extracted paraphrases is fairly similar across most languages, the cross-lingual models do not perform equally well across all languages, as indicated by the precision and recall of these models.
Differences between languages are present also for stricter decision thresholds.
A further cause of errors is translations. While all datasets are translated from the same English originals, the translations are not necessarily direct and unambiguous. For example, ``I have a problem with a mole'' can mean a problem due to skin growth or caused by an animal, and can get different translations due to the assumed meaning.
To illustrate the variance of the extracted paraphrases in the cross-lingual setting, we counted the overlapping paraphrases for the XLM-R-large model with the argmax threshold. Across all $15$ languages, there are a total of $898$ unique paraphrases, of which $47$ are common across all languages, and $330$ are common across at least $10$ languages.

The results of manual validation show that the paraphrases extracted using a cross-lingual approach are also of good quality. For the two settings we validated (German and French), we found that approximately $80\%$ of the extracted paraphrases are correct.

In addition to cross-lingual approach, we tested two common translation baselines ``translate-train''  and ``translate-test''  \cite{conneau-etal-2018-xnli}.
The ``translate-train'' generally extracts more paraphrases than the cross-lingual models, while the reverse is true for the ``translate-test'' approach.
We hypothesize this is due to the noise introduced by machine translation.
In ``translate-train'', the training sequences are translated into the target language.
The low quality of some translations causes that the models learn noisy patterns and extract more paraphrases from the validation and test set.
In ``translate-test'', the situation is reversed. The model learns from higher-quality English instances, while the machine-translated validation and test instances are less coherent and less likely classified as paraphrases.
For example, the German sentence ``Was super ist am Leben auf dem Land, ist dass man sich nicht über solche Dinge ärgern muss.'' gets incorrectly translated into ``What's great about life in the country is that you don't have to \emph{tease} about such things.'' (instead of ``be upset''), which decreases the entailment probability in this instance.
Due to the sources of noise that are hard to account for, we warn against the sole use of translation baselines in XNLI. The cross-lingual models present a strong alternative and may be used alone or in an ensemble mode. 

\begin{table*}[ht]
\caption{The number of filtered paraphrases (\#), precision (P), recall (R) and manually estimated precision (\faHandPaperO P) in QQP and MRPC datasets using three different decision thresholds for the non-paraphrase label. Below the dataset name we show the number of paraphrase instances in the dataset.} 
\centering
\resizebox{0.75 \linewidth}{!}{
\begin{tabular}{cl@{\hskip 30pt}cccc@{\hskip 30pt}cccc}
\toprule
dataset & decision & \multicolumn{4}{c}{roberta-base} & \multicolumn{4}c{roberta-large} \\
(size) & threshold & \# & P & R & \faHandPaperO P & \# & P & R & \faHandPaperO P \\
\midrule
\multirow{3}{*}{\makecell{QQP \\ (\num{149263})}}
& argmax      & $13471$ & $0.934$ & $0.834$ & $0.940$ & $13426$ & $0.937$ & $0.884$ & $0.910$ \\
& T=$0.75$    & $4084$ & $0.978$ & $0.727$ & $0.940$ & $5865$ & $0.967$ & $0.821$ & $0.920$ \\
& T=$0.90$    & $1285$ & $0.991$ & $0.622$ & $0.950$ & $2363$ & $0.985$ & $0.748$ & $0.960$ \\
\midrule
\multirow{3}{*}{\makecell{MRPC \\ (\num{3900})}}
& argmax    & $230$ & $0.854$ & $0.718$ & $0.900$ & $213$ & $0.832$ & $0.796$ & $0.860$ \\
& T=$0.75$  & $157$ & $0.882$ & $0.649$ & $0.920$ & $148$ & $0.859$ & $0.746$ & $0.950$ \\
& T=$0.90$  & $83$ & $0.913$ & $0.542$ & $0.960$ & $100$ & $0.880$ & $0.670$ & $0.940$ \\
\bottomrule
\end{tabular}
}
\label{tab:results-qqp-mrpc}
\end{table*}

\subsection{Filtering existing paraphrases}
\label{sec:filtering-results}
Using the methodology described in \Cref{sec:paraphrase-filtering}, we cleaned two existing paraphrasing datasets (QQP and MRPC) described in \Cref{sec:datasets}. Table \ref{tab:results-qqp-mrpc} shows the number of filtered-out paraphrases from QQP and MRPC datasets and a manual estimate of their precision. As in our paraphrase extraction experiments, we report the results using three decision boundaries, but now the thresholds apply to probabilities of non-paraphrase labels.

Our method removes a relatively high amount of false paraphrases for both datasets, considering both datasets are well-known and popular paraphrase identification datasets. This confirms recent findings in other machine learning areas \cite{northcutt2021pervasive}, where analyses have shown relatively high noise levels with possibly detrimental consequences. 

The false paraphrases in QQP are often a consequence of using questions with the same intent but not necessarily identical concepts, as mentioned in \Cref{sec:paraphrase-filtering}. In MRPC, the false paraphrases are due to relaxed annotation guidelines of what is to be considered a paraphrase. 
\citet{dolan-brockett-2005-mrpc} mention that their dataset intentionally contains sentence pairs that semantically differ to an extent, a choice made to increase the diversity of paraphrases. If a dataset user does not agree with this loose approach or wishes to treat such instances differently, our method can automatically find at least some of them.

Similarly to \Cref{tab:results-snli-mnli}, the results in \Cref{tab:results-qqp-mrpc} also show that a more strict decision boundary leads to less filtered-out paraphrases. However, these are more likely to be false paraphrases.

The results of manual validation show that the filtered-out paraphrases are, in most cases, actual non-paraphrases. The filtered-out paraphrases typically contain more than $90\%$ of actual non-paraphrases.
Qualitative analysis of the filtered-out instances shows that they are often similarly structured, with one of the sequences containing additional information that is not present in the other sequence, making the pair semantically non-equivalent.
Among the erroneously filtered-out paraphrases, we find that some are filtered due to too strict NLI models. For example, when processing the pair ``What are CoCo bonds?'' and ``What is a coco bond?'', the model decides that the two sentences are not paraphrases, likely because one is talking about bonds (plural) and the other is talking about a bond (singular). In a semantic sense, this pair shall be declared a paraphrase.

\section{Conclusion}
\label{sec:conclusions}
We have proposed a novel paraphrases extraction approach based on the similarity between NLI and paraphrasing. Results show that the proposed methodology can effectively extract paraphrases from NLI datasets both in the monolingual and cross-lingual setting. Furthermore, the proposed filtering demonstrates a surprisingly large amount of noise in the existing paraphrasing datasets.
In summary, the proposed method enables the reuse of NLI resources and provides additional quality assurances for paraphrasing. 

In further work, we plan to extend the analysis of extracted paraphrases and test their use in downstream tasks. For example, different decision thresholds and different qualities of paraphrases might be helpful in different applications -- to favour either logical equivalence or variability of sentences.

Additionally, there is a potential to leverage further connections between NLI and other tasks. For example, text summarization instances could possibly be obtained from NLI datasets by extracting sequence pairs where entailment holds in one direction, while the instances are neutral in the other direction. Another more difficult task would be to extract NLI instances from existing paraphrasing datasets, i.e. going in the reverse direction of the presented one.

 \subsection*{Acknowledgements}
 
For the evaluation of French paraphrases we are grateful to Jožica Robnik-Šikonja.
The work was partially supported by the Slovenian Research Agency (ARRS) through the core research programme P6-0411, projects  J6-2581 and J7-3159, as well as the young researcher grant. This paper is supported by European Union's Horizon 2020 research and innovation programme under grant agreement No 825153, project EMBEDDIA (Cross-Lingual Embeddings for Less-Represented Languages in European News Media).

\clearpage
\bibliographystyle{plainnat}  
\bibliography{anthology,custom} 

\end{document}